# Research on Joint Representation Learning Methods for Entity Neighborhood Information and Description Information

Le Xiao [1], Xin Shan [1], Yuhua Wang[1] and Miaolei Deng [2]

[1] School of Information Science and Engineering, Henan University of Technology, zhengzhou, 450001, China.
[2]Graduate School, Henan University of Technology,zhengzhou,450001,China.
dml_1978@foxmail.com

**Abstract.** To address the issue of poor embedding performance in the knowledge graph of a programming design course, a joint representation learning model that combines entity neighborhood information and description information is proposed. Firstly, a graph attention network is employed to obtain the features of entity neighboring nodes, incorporating relationship features to enrich the structural information. Next, the BERT-WWM model is utilized in conjunction with attention mechanisms to obtain the representation of entity description information. Finally, the final entity vector representation is obtained by combining the vector representations of entity neighborhood information and description information. Experimental results demonstrate that the proposed model achieves favorable performance on the knowledge graph dataset of the programming design course, outperforming other baseline models.

## 1 Introduction

Knowledge graph representation learning is a method that transforms entities and relationships in a knowledge graph into low-dimensional vectors, enabling efficient computation of complex semantic associations [1]. It serves as the foundation for downstream tasks such as knowledge reasoning and knowledge base construction. In the knowledge graph of a programming design course, a target entity is connected to other relevant entities through relationships, and the relationships and connected entities starting from the target entity are referred to as its structural neighborhood. Different entities within the neighborhood have varying degrees of importance to the target entity, and entities typically possess rich description information.
As illustrated in Fig.1, the target entity "Array" is connected to neighboring entities such as "Memory" and "One-dimensional Array" through relationships like "require" and "include". Additionally, it is accompanied by the description information "Definition of an array". Traditional representation learning methods for course knowledge graphs typically focus on embedding individual entities without considering their neighborhood information and description information, leading to suboptimal embedding performance.

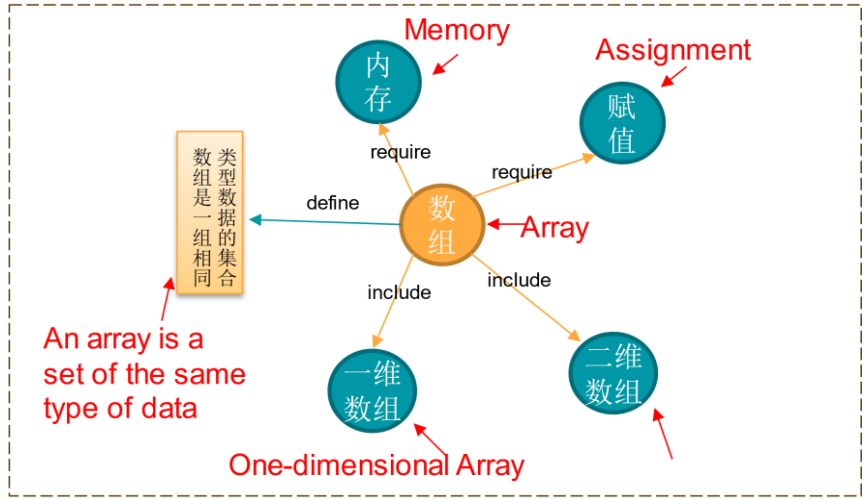

**Fig.1.** Sample Knowledge Graph of Programming Course

This paper proposes a representation learning model, named NDRL (A Representation Learning Model for Joint Entity Neighborhood Information and Description Information), based on the characteristics of the programming design course knowledge graph. The model aims to effectively integrate entity neighborhood information and description information, utilizing the information within the knowledge graph to obtain high-quality embedding representations. This model plays a significant role in subsequent tasks such as knowledge reasoning [2,3], completion [4,5], and applications [6,7] based on the course knowledge graph.

## 2 Related Work

Knowledge graphs are often represented symbolically, which can lead to issues such as low computational efficiency and data sparsity [8]. With the development and application of deep learning, there is a growing desire for more simple and efficient representations of knowledge graphs. This has given rise to knowledge graph representation learning methods, which aim to map the elements of knowledge graphs, including entities and relationships, into a continuous low-dimensional vector space. The goal is to learn vector representations for each element in the vector space, thereby mapping the triplets from a high-dimensional one-hot vector space to a continuous low-dimensional dense real-valued vector space [9]. This approach addresses the problem of data sparsity in knowledge bases and enables efficient computation.

Among the existing research, translation models are the most representative and classic methods [10]. The fundamental idea behind translation models is to map entities and relationships into a shared vector space, where the semantic information of entities and relationships can be represented by the similarity between vectors. However, these methods typically independently learn the structural features of each triplet, without incorporating the semantic information present in the knowledge graph.

To address this issue, several knowledge graph representation learning methods based on semantic information have been proposed in recent years. KG-BERT[11] represent entities and relations as their name/description textual sequences and turn the knowledge graph completion problem into a sequence classification problem. The RotatE[12] model defines each relation as a rotation from the source entity to the target entity in the complex vector space, which is able to model and infer various relation patterns. The methods based on Graph Convolutional Neural Networks (GCN) [13] are one of the most commonly used approaches. GCN learns the semantic relationships between entities and relationships by performing convolutions on the graph, thereby integrating this information into vector representations.

In recent years, researchers have proposed improved graph neural network models, such as Graph Attention Network (GAT) [14]. GAT utilizes attention mechanisms to learn the interaction between neighbor nodes. Compared to GCN, GAT not only captures the complex relationships between nodes more comprehensively but also exhibits better interpretability and generalizability.

In addition to the structural information of the triplets themselves, knowledge graphs often contain rich additional information such as entity descriptions and attribute information. Xie et al. [15] incorporated entity description information from the knowledge graph into knowledge graph representation learning and proposed the DKRL model. The model used both convolutional neural networks and continuous bag-of-words models to encode entity description information. It leveraged both factual triplets and entity description information for learning and achieved good inference performance. However, since these representations did not include the entire semantic information of entity descriptions, there might be some loss of semantic information. Additionally, in many large-scale knowledge graphs, there is a lack of entity descriptions for many entities. To address this, Wang et al. [16] introduced an external text corpus and used the semantic structures of entities in the text corpus as part of the entity representation, further improving the accuracy of knowledge inference in cases of missing entity descriptions. Reference [17] proposed a rule-guided joint embedding learning model for knowledge graphs. It utilized graph convolutional networks to fuse context information and textual information into the embedding representation of entities and relationships, further enhancing the representation capability of entities and relationships. Inspired by translation-based graph embeddings designed for structural learning, Wang et al.[18]apply a concatenated text encoder. Then, a scoring module is proposed based on these two representations, in which two parallel scoring strategies are used to learn contextual and structural knowledge.

In this work, we address the following issues based on existing work [15,19]. The following research and improvements have been conducted:

1. A representation learning model is proposed that jointly considers entity neighborhood information and description information. Improved GAT and BERT models are used to obtain vector representations of entity neighborhood information and description information, respectively, to fully utilize the hidden complex entity relationship feature vectors in knowledge graph triplets and represent the target entity.

2. Joint learning is performed on the vector representations of entity neighborhood information and description information, training both types of representations in the same continuous low-dimensional vector space to better utilize the two different types of information.

3. For entities that already have rich neighborhood information, to avoid noise interference caused by the addition of description information, the concept of "entity structure richness" is defined. Based on the magnitude of entity structure richness, different representation learning methods are selected to obtain the optimal vector representation. Experiments on a dataset of programming course knowledge graph demonstrate that the proposed model outperforms other baseline models.

## 3. Our Model

In the entity neighborhood information representation module, we employ an improved Graph Attention Network (GAT) model to obtain the embedding representation. In the entity description information module, we use the BERT-WWM model and attention mechanism to obtain the corresponding embedding representation. Different approaches are selected to obtain the final entity vector representation based on the entity structure richness. The model framework is illustrated in Fig.2.

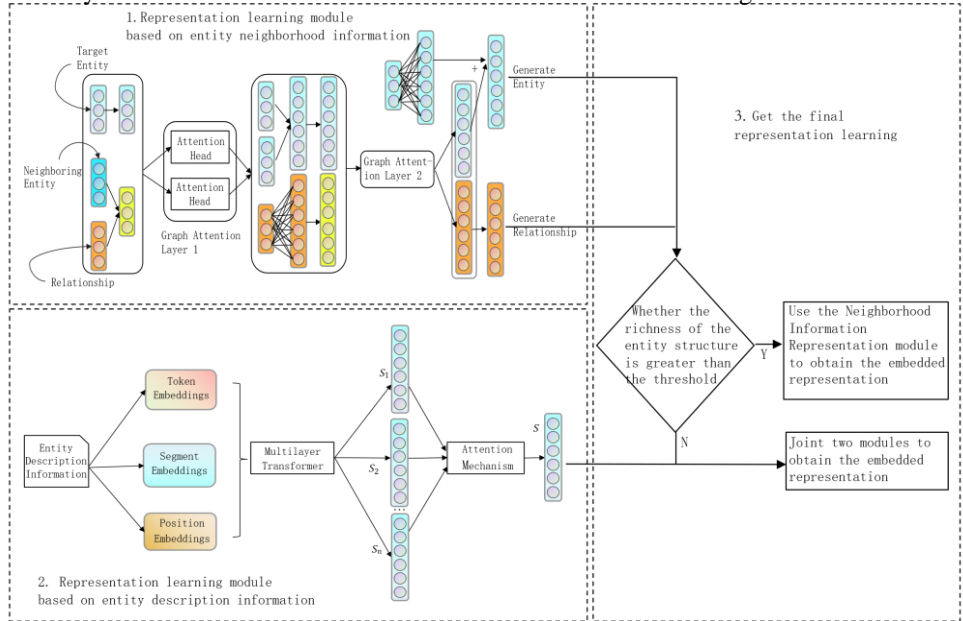

**Fig.2.** Representation learning model with joint entity neighborhood information and description information

### 3.1 Representation Learning Based on Entity Neighborhood Information

Traditional Graph Attention Network (GAT) models learn the weights of neighboring nodes to perform weighted summation of their features, but they do not consider the importance of relationships for entity representation. To incorporate relationships as important information during training and combine them with the structural features of triplets in the knowledge graph, we enhance the GAT model by adding relationships as significant information to the graph attention mechanism. Specifically, to apply the attention mechanism to the target node and its neighboring nodes, we first compute weighted sums of entities and relationships in the neighborhood. The construction methods for the target node and its neighboring nodes are defined as equations (1) and (2):

$$h_i = t_s \tag{1}$$

$$h_j = \rho h_s + (1 - \rho) r_s \tag{2}$$

Among them, $h_s$, $t_s$ and $r_s$ represent the initial vector representations of the head entity, tail entity, and relationship, respectively. The weight parameter $\rho \in (0,1)$ is used to adjust the proportion of the relationship vector compared to the entity vector when constructing neighboring nodes. This allows both the entity and the relationship of each triplet to participate in the computation of the graph attention model. To calculate the influence weight of $h_j$ on the target node $h_i$ we define their attention value $v_{ij}$ as shown in equation (3):

$$v_{ij} = a(Wh_i, Wh_j) \tag{3}$$

Whereas: parameter $W$ represents the projection matrix, and the attention mechanism a is a single-layer feed-forward neural network. Expanding equation (3) yields the specific calculation formula:

$$v_{ij} = LeakyRelu(z^T[Wh_i \parallel Wh_j]) \tag{4}$$

After multiplying the projection matrix with the feature vectors and concatenating them together, a linear transformation is applied using the weight vector $z$. Then, a nonlinear activation is performed using the LeakyReLU function. Finally, the Softmax function is applied to normalize the attention values between each node and all its neighboring nodes. The normalized attention weights serve as the final attention coefficients, as shown in equation (5):

$$\alpha_{ij} = Softmax_j(v_{ij}) = \frac{\exp(v_{ij})}{\sum_{k \in N_i} \exp(v_{ik})} \tag{5}$$

Where $N_i$ represents the neighboring nodes of the target node $h_i$, which consists of the entities $h_s$ adjacent to the target node $t_s$ and the relations $r_s$ between them as defined in equation (2). The attention coefficients calculated are then weighted and summed up as shown in equation (6):

$$h_i^{'} = \sigma\left(\sum_{j \in N_i} \alpha_{ij} Wh_j\right) \tag{6}$$

Where $h_i^{'}$ represents the new feature vector for each node $i$ based on the output of GAT, which integrates the neighborhood information of entities in the knowledge graph. The function $\sigma$ is the activation function, and the output of the target node is related to the feature vectors of all neighboring nodes. To enable the model to learn the features of neighboring nodes more stably, a multi-head attention mechanism is used to obtain different features for integration. To prevent overfitting, the vectors obtained from K-independent attention mechanisms are concatenated. The specific representation is given by equation (7):

$$h_i^{'} = \parallel_{k=1}^{K} \sigma\left(\sum_{j \in N_i} \alpha_{ij}^k W^k h_j\right) \tag{7}$$

In the last layer of the graph attention model, the obtained vector representations are averaged instead of concatenated. This can be expressed by equation (8):

$$h_i^{'} = \sigma\left(\frac{1}{K} \sum_{j \in N_i} \alpha_{ij}^k W^k h_j\right) \tag{8}$$

To ensure that relation vector representations have the same output dimension as entity vector transformations, they will share the output dimension. Following a graph attention calculation, the relation vectors undergo a linear transformation, as depicted in equation (9):

$$R^{'} = RW^R \tag{9}$$

The input set of relation vectors is denoted as R, and $W^R \in R^{T \times T^{'}}$ represents the linear transformation matrix. Here, T corresponds to the dimension of the original vectors, and T^' represents the dimension after transformation. However, in the process of obtaining new entity vector representations, there is a potential loss of the original structural features. To tackle this issue, the initial entity vectors undergo a linear transformation and are then added to the final entity representations in the following manner:

$$E^{'} = EW^E + E^f \tag{10}$$

$W^E \in R^{T^i \times T^f}$, The parameter $T^i$ represents the initial dimension of entity vectors, while $T^f$ represents the final dimension. $E$ represents the set of initial input entity vectors, and $E^f$ denotes the set of entity vector representations learned through GAT.

**3.2 Representation Learning Based on Entity Description Information**

In this paper, the BERT model is introduced to represent the complete entity description information. The entity description information serves as the direct input to the BERT model, minimizing information loss and capturing the full semantic representation of the entity description. The model employs a multi-layer Transformer structure, which captures bidirectional relationships within sentences. However, since the BERT model masks individual characters during training and does not consider Chinese word segmentation conventions, this paper utilizes the BERT-WWM model, an upgraded version of BERT specifically optimized for Chinese tasks. It incorporates the whole-word masking technique, allowing for better handling of the complex language structure in Chinese. As shown in Table 1, through tokenization, the input text "数组经常被用作实际参数" (Arrays are often used as actual parameters) is segmented into several words, such as "数组" (arrays), "经常" (often), "被" (are), "用作" (used as), "实际" (actual), and "参数" (parameters). Traditional BERT masking randomly selects words for masking, for example, replacing "组" (group) and "参" (participation) with the [MASK] token. In contrast, according to the BERT-WWM model, the "数" (number) in "数组" (arrays) and the "数" in "参数" (parameters) would also be replaced by the [MASK] token. This enhancement aims to improve the model's performance.

**Table 1** Example table of whole word MASK

| Input text | 数组经常被用作实际参数 |
|---|---|
| word segentation | 数组 经常 被 用作 实际 参数 |
| BERT masking mechanism | 数[MASK] 经常 被 用作 实际 [MASK]数 |
| BERT-WWM masking mechanism | [MASK][MASK] 经常 被 用作 实际 [MASK][MASK] |

Firstly, the entity description information is transformed into word embedding, segmentation embedding, and location embedding; then it is vector stitched as the input of the BERT-WWM model, and the sentence vector $S_i$ ($i=1,2,…,n$) of this entity description information is obtained by multi-layer Transformer structure, which is represented as the sentence vector of the $i$-th sentence.

After that, using the vector representation $h_i^{'}$ of the target entity obtained in the entity neighborhood representation module, the influence weight of each sentence vector $S_i$ of the description information of the target entity is calculated in the same way as Eqs. (3)-(8), and the attention weight of $S_i$ depends on the correlation between $S_i$ and $h_i^{'}$, from which the attention weight distribution of each sentence vector is calculated, and the weighted aggregation of each sentence vector representation is obtained to obtain the entity vector representation $S$ of descriptive information.

### 3.3 Obtaining the final embedding representation

After obtaining the vector representations of the above two modules separately, this paper takes two approaches to the two vector representations, one is to perform a joint representation of the two, by combining the triadic structural information of the knowledge graph and entity descriptions for the training of the model in an integrated manner. The vector representations of entities and relations are learned in the same continuous low-dimensional vector space. The energy function of the synthesis is defined as shown in equation (11)

$$d = d_g + d_w \qquad (11)$$

Where $d_g = \| h_g + r - t_g \|_2$ is the GAT-based energy function and the $h_g$ and $t_g$ sub-tables are the GAT-based representations of the head entity and tail entity; $d_w$ is the energy function based on the description information. To achieve unity between the two in the same vector space, the relational representation r in GAT is used in training, and $d_w$ is defined as in equation (12)

$$d_w = d_{ww} + d_{wg} + d_{gw} \qquad (12)$$

where $d_{ww} = \| h_w + r - t_w \|_2$, $h_w$ and $t_w$ are the description information-based representations of the head entities and tail entities, respectively. In $d_{wg}$ and $d_{gw}$, one of the head and tail entities is represented using a vector based on entity description; the other is represented using a vector obtained from GAT training as $d_{wg} = \| h_w + r - t_g \|_2$ and $d_{gw} = \| h_g + r - t_w \|_2$, respectively. The two types of representations are jointly trained in the above way to obtain the final vector representation.

However, after our experiments, we found that when an entity has more neighboring entities, joint representation learning is not yet as effective as using only the neighborhood information module. This is because when the entity has richer neigh-bors, these neighbors are already able to provide enough information to the entity, and then joined with the description information at this time will bring some noise instead, which affects the embedding effect. Therefore, this leads to the second treatment, which is to use only the vector representation obtained in the neighborhood information representation module as the final vector representation, and therefore, this involves the question of a threshold, i.e., which entity is represented by the joint representation and which entity is represented by the neighborhood information representation module. Therefore, we define a concept called "entity structure richness" to measure the size of entity neighborhood information, and select different representation learning methods according to the size of entity structure richness, which is defined as shown in equation (13).

$$N(e) = n_e + kn_{N_e} \qquad (13)$$

where $n_e$ is the entity degree, $n_{N_e}$ is the degree of entity neighbor nodes, and $k$ is the hyperparameter in the range of 0-1.

### 3.4 Loss function

Based on the above computational analysis, the loss function is further constructed. A boundary-based optimization method is defined and used as the training objective to optimize this model by minimizing the loss function L. Both vector representations in the model use this loss function.

$$L = \sum_{(h,r,t)\in T} \sum_{(h',r',t')\in T'} max\left(\gamma + d(h,r,t) - d(h',r,t'), 0\right) \qquad (14)$$

where γ is the boundary parameter measuring the correct and incorrect triples. $T$ is the set of positive examples consisting of the correct triples $(h, r, t)$ and $T'$ is the set of negative examples consisting of the incorrect triples $(h', r, t')$, and $T'$ is defined as shown in Equation (15):

$$T' = \{(h',r,t)|h' \in \varepsilon\} \cup \{(h,r,t'|t' \in \varepsilon)\} \qquad (15)$$

$T'$ in Equation (15) is obtained by randomly replacing the head entity, tail entity, or relationship in the set of positive examples to obtain the corresponding set of negative examples. During the training of the model, the optimization operation is performed using stochastic gradient descent to minimize the value of its loss function.

## 4 Experiment

### 4.1 Dataset

At present, there is no public authoritative knowledge graph of programming classes in the field of knowledge mapping representation learning. In this paper, the course knowledge obtained from the "C Programming" textbook, teacher's courseware, Baidu encyclopedia, etc. are used as data sources, and the course knowledge points are used as entities, the description information of knowledge points as attributes and three entity relationships between knowledge points and knowledge points are set, namely, " include": indicates the inclusion relationship between entities of similar knowledge points, "require": indicates the logical dependency relationship between knowledge points, "relate": indicates the relates": denotes the related relationship between knowledge points. Thus, the programming course knowledge graph dataset (PDCKG) was constructed. A total of 2685 entities and 9869 triples were obtained, and the training set, validation set, and test set were divided according to the ratio of 7:1.5:1.5.

**4.2 Parameter settings**

In obtaining the GAT-based representation, the TransE training vector based on Xavier initialization is used as the initialized structural feature vector representation of entities and relationships. The alpha parameter of LeakyReLU is set to 0. 2. In order to prevent overfitting of the model, L2 regularization is used; in order to obtain the BERT-WWM-based representation of entity description information, the alpha parameter of LeakyReLU is set to 0.2. The hyperparameter k in the entity richness expression is set to 0.5, and the entity structure richness threshold is set to 12, i.e., when the entity structure richness is not less than 12, the vector representation obtained based on the neighborhood information module is used as the final representation, and vice versa, the vector representation obtained from the joint representation is used as the final representation. Let the learning rate λ be 0.004, the boundary value γ be 1.0, and the size of the batch be 512 during the model training.

**4.3 Link prediction task experiment**

Link prediction aims to test the inferential prediction ability of the model. For a given correct triple $(h,r,t)$, after the missing head entity h or tail entity $t$, the head and tail entities are randomly selected in the original entity set to complete the set, and for the missing positions, the scores of the reconstituted triples are calculated by the model, and then sorted in ascending order, and the ranking of the final correct triple is recorded. That is, the tail entity t in the missing triple $(h,r,)$ or the head entity $h$ in the missing triple $(,r,t)$ is predicted. Mean Rank (MR), Mean Reciprocal Rank (MRR), Hits@1, and Hits@10 are standard evaluation measures for the dataset and are evaluated in our experiments.

In addition, there is a problem that when constructing negative samples, the new triple formed after replacing the head entity or tail entity may already exist in the knowledge graph, which may interfere with the actual ranking of the correct triple and have some influence on the evaluation results. Therefore, in this paper, the link prediction experiments are divided into "raw" and "filter" according to whether to filter the existing triples or not. The experimental results of each model on PDCKG are shown in Table 2.

**Table 2.** Effectiveness of link prediction for each model

| Model | hits@1/% | | hits@10/% | | MR | | MRR | |
|---|---|---|---|---|---|---|---|---|
| | **Filter** | **Raw** | **Filter** | **Raw** | **Filter** | **Raw** | **Filter** | **Raw** |
| TransE | 8.39 | 5.85 | 32.34 | 20.54 | 556.98 | 804.51 | 0.126 | 0.091 |
| DKRL | 10.11 | 7.64 | 35.79 | 22.66 | 387.25 | 519.55 | 0.185 | 0.156 |
| R-GCN | 10.41 | 7.28 | 28.09 | 17.85 | 679.99 | 906.79 | 0.173 | 0.148 |
| RotatE | 25.68 | 20.35 | 51.84 | 40.12 | 201.36 | 341.66 | 0.346 | 0.296 |
| KG-BERT | 22.07 | 17.26 | 48.25 | 39.16 | 190.58 | 301.52 | 0.365 | 0.287 |
| KBGAT | 26.87 | 20.96 | 53.25 | 44.15 | 185.62 | 310.19 | 0.304 | 0.245 |
| StAR | 21.06 | 17.34 | 44.38 | 37.91 | 180.55 | 292.36 | 0.283 | 0.251 |
| **NDRL（Ours）** | **28.64** | **21.58** | **64.13** | **50.39** | **105.63** | **198.68** | **0.387** | **0.332** |

It can be seen that on the filtered dataset, each model performs better than the original dataset, which illustrates the necessity of performing filtering operations, and NDRL outperforms other comparable models in all indexes. Compared with the DKRL model, our model uses BERT to represent the entity description information, which can minimize the loss of information and obtain the entity description as much as possible. Compared with the R-GCN model, our model uses the method of constructing neighbor nodes by combining entities and relations so that the target node combines more information about entities and relations in the neighborhood, which improves the inference ability of the model. Compared with KBGAT, because the structure-based model only considers the structured information of the triad, when the corresponding information is missing, it will not be able to make predictions, the entity description information can be used as a favorable supplement to the structure-based model, thus improving the ability of knowledge graph representation learning and the performance of prediction, and therefore, the model has a richer representation capability.

**4.4 Ablation Experiment**

To further verify the effectiveness of the model in this paper, some modules in the model in this paper are removed and compared with the model in this paper, Hits@1, Hits@10, and MR are used as evaluation indexes, and the comparison models are as follows:

1. (NDRL-r): In the entity neighborhood information representation module, the traditional GAT is used to obtain the entity vector representation without adding the relationship into the model.

2. (NDRL-a): In the entity description information representation module, after obtaining the sentence vectors, the attention mechanism is not used to obtain the final vector representation, and the traditional averaging summation method is used.

3. (NDRL-s): In the module of obtaining the final vector representation, the final result is obtained directly by using the joint representation learning method without doing the discriminant of entity structural richness.

The experimental results are shown in Table 3.

**Table 3.** Ablation experiment

| Model | hits@1/% | | hits@10/% | | MR | | MRR | |
|---|---|---|---|---|---|---|---|---|
| | Filter | Raw | Filter | Raw | Filter | Raw | Filter | Raw |
| NDRL-r | 25.58 | 18.69 | 58.65 | 44.98 | 150.64 | 272.36 | 0.364 | 0.305 |
| NDRL-a | 26.21 | 19.36 | 60.36 | 48.68 | 125.65 | 223.85 | 0.371 | 0.311 |
| NDRL-s | 25.28 | 19.10 | 57.01 | 45.54 | 161.27 | 279.35 | 0.349 | 0.296 |
| **NDRL** | **28.64** | **21.58** | **64.13** | **50.39** | **105.63** | **198.68** | **0.387** | **0.332** |

As can be seen from the table, for each metric, removing any of the modules from the model makes the performance of the model decrease compared to the NDRL model, so the effect of all modules is positive.

**4.5 Error Analysis**

During the experiment, we found that there are many ORC (one-relation-circle) structures on the PDCKG dataset, i.e., structures composed of some special relations such as symmetry and propagation relations, etc. After statistics, the ORC structure data in PDCKG accounted for 15.38% of the total data. And the translation model based on TransE cannot handle such ORC structures. For example, for the symmetric relation $r$, there exist $(h,r,t) \in G$ and $(t,r,h) \in G$. In TransE, the corresponding vectors then satisfy $h + r \approx t$ and $t + r \approx h$, which can be mathematically introduced: $h \approx t$ and $r \approx 0$. This leads to the inability to model, which has a certain impact on the experimental effect. In the subsequent work, other more effective modeling approaches will be explored to deal with the ORC structure present in the data and make the model obtain better performance.

11## 5 Summary

In this paper, we proposed a representation learning model, NDRL, for the union of entity domain information and description information, which integrates the entity neighborhood information and entity description information in the knowledge graph of programming courses. First, a combined representation of the relationships of entities and neighboring entities using the GAT-based representation learning model is used to obtain the corresponding neighborhood information representation; then, the entity description information is encoded and represented by the BERT-WRM model to obtain the entity description information representation corresponding to the entities; finally, it is integrated into a joint model for joint training and learning, and the experimental on the PDCKG dataset The results show that the NDRL model proposed in this paper can improve the performance of link prediction and triad classification tasks well compared with other benchmark models.

In our future work, we will further investigate the representation learning method for the knowledge graph of programming courses and hope to improve it in the following 2 aspects: 1) The model in this paper focuses on using the neighborhood information and entity description information of entities, but it has not been utilized for the knowledge graph such as category information and other knowledge base information, so the joint knowledge representation learning method of multiple sources is still future research and improvement.2) For the ORC structures existing in the knowledge graph of programming courses, we will further explore other more effective modeling approaches to deal with such structures so that the model can obtain better performance.